\documentclass[10pt,twocolumn,letterpaper]{article}

\usepackage{cvpr}
\usepackage{multirow}
\usepackage{float}
\usepackage{graphicx}
\usepackage{subcaption}
\usepackage{siunitx}
\usepackage{tabularx}
\usepackage{amsmath}
\usepackage{booktabs}
\usepackage{xcolor} 
\usepackage{bibunits}
\usepackage{cuted}
\usepackage{capt-of}
\usepackage[absolute]{textpos}
\defaultbibliographystyle{ieee_fullname}

\definecolor{cvprblue}{rgb}{0.21,0.49,0.74}
\usepackage[breaklinks,colorlinks,allcolors=cvprblue]{hyperref}

\def\paperID{28245} 
\def\confName{CVPR}
\def\confYear{2026}

\title{DETACH : Decomposed Spatio-Temporal Alignment for Exocentric Video and Ambient Sensors with Staged Learning}

\author{Junho Yoon \quad Jaemo Jung \quad Hyunju Kim \quad Dongman Lee\\
KAIST\\
{\tt\small \{vpdtlrdl, gosfl4760, iplay93, dlee\}@kaist.ac.kr}
}

\begin{document}

\begin{bibunit}[ieeenat_fullname]
    \maketitle
    \begin{abstract}
Aligning egocentric video with wearable sensors have shown promise for human action recognition, but face practical limitations in user discomfort, privacy concerns, and scalability. We explore exocentric video with ambient sensors as a non-intrusive, scalable alternative. While prior egocentric-wearable works predominantly adopt Global Alignment by encoding entire sequences into unified representations, this approach fails in exocentric-ambient settings due to two problems: (P1) inability to capture local details such as subtle motions, and (P2) over-reliance on modality-invariant temporal patterns, causing misalignment between actions sharing similar temporal patterns with different spatio-semantic contexts. To resolve these problems, we propose DETACH, a decomposed spatio-temporal framework. This explicit decomposition preserves local details, while our novel sensor-spatial features discovered via online clustering provide semantic grounding for context-aware alignment. To align the decomposed features, our two-stage approach establishes spatial correspondence through mutual supervision, then performs temporal alignment via a spatial-temporal weighted contrastive loss that adaptively handles easy negatives, hard negatives, and false negatives. Comprehensive experiments with downstream tasks on Opportunity++ and HWU-USP datasets demonstrate substantial improvements over adapted egocentric-wearable baselines.
\end{abstract}
    \section{Introduction}
\label{sec:introdcution}

Human action recognition (HAR) has been extensively studied using video \cite{li2025challenges, min2021integrating} and sensor modalities \cite{ordonez2016deep, yao2017deepsense}, with each offering unique strengths as video provides rich visual context \cite{kong2022human} and sensors offer precise temporal cues \cite{ni2024survey}. 
Recently, multimodal approaches aligning egocentric video with wearable sensors \cite{moon2022imu2clip, das2025primus, chen2025comodo, zhang2024masked} have demonstrated promising results in personal action monitoring by leveraging their complementarity.
However, such approaches face practical limitations in real-world deployment, as they introduce ongoing user burden, expose sensitive personal data, and offer limited scalability across multiple users \cite{serpush2022wearable, zhang2022deep}.

\begin{figure}[tbp]  
    \centering
    \includegraphics[width=0.47\textwidth]{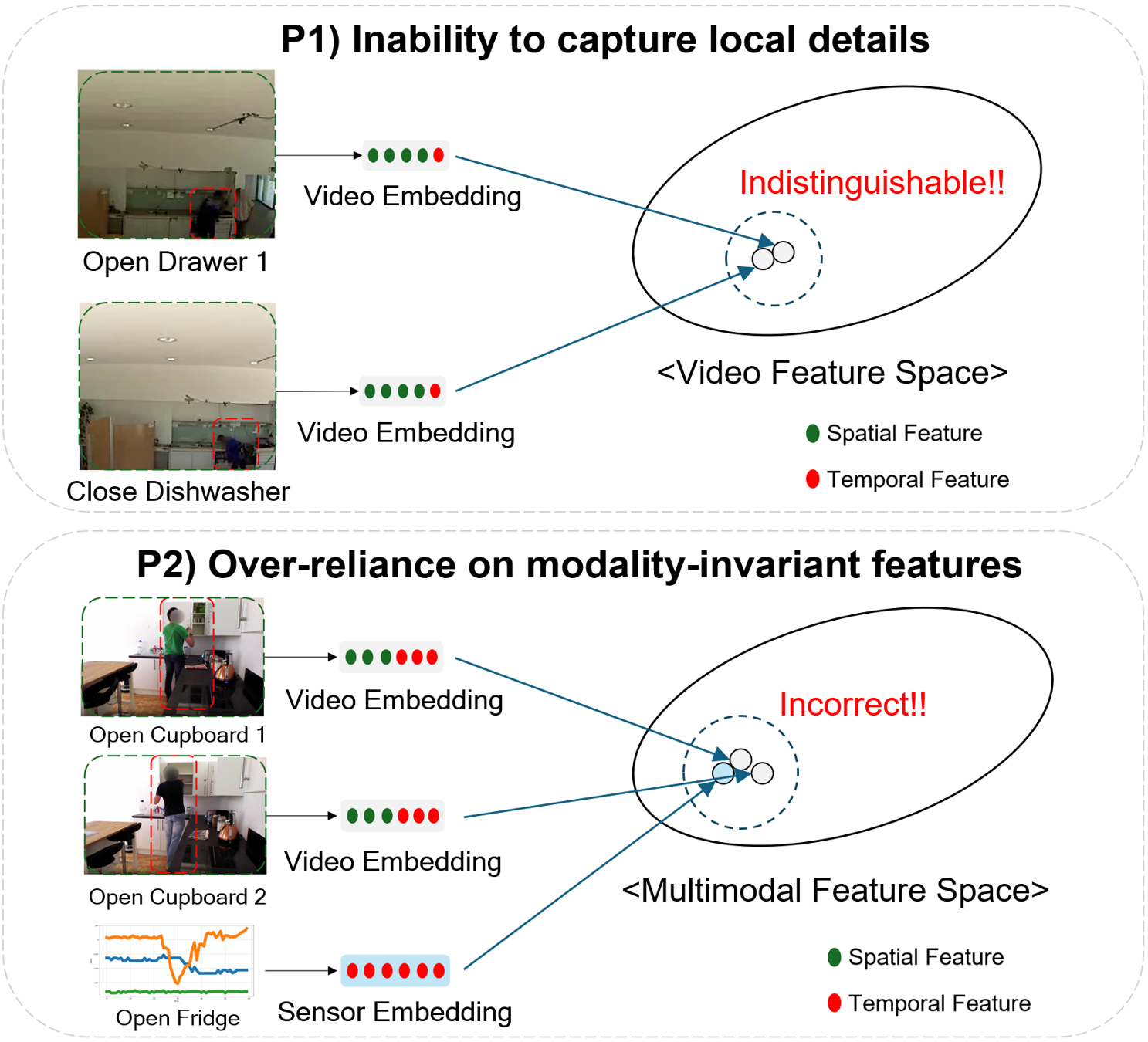}
    \vspace{-0.2cm}
    \caption{{\textbf{Limitations of Global Alignment-based approaches.}
    (a) Inability to capture local details: subtle motions become indistinguishable in video feature space.
    (b) Over-reliance on modality-invariant features: actions incorrectly aligned due to shared temporal patterns, despite different spatio-semantic contexts.}}
    \label{fig:globalAlignment}
\end{figure}

Integrating exocentric video with ambient sensors offers a promising direction for addressing the limitations. 
Unlike egocentric–wearable setups, exocentric–ambient configurations employ statically installed cameras \cite{kong2022human} and object-embedded sensors \cite{nabiei2015object}, enabling non-intrusive and scalable monitoring \cite{van2008accurate, rashidi2012survey}.
However, despite this potential, no existing work has explored exocentric–ambient alignment, and the \textit{Global Alignment} strategy commonly used in the egocentric–wearable domain is unsuitable to this setting. 
Global Alignment strategy encodes entire video clips and sensor windows into holistic representations, which are then aligned through contrastive or distillation objectives \cite{liu2019use, gabeur2020multi}. 
While effective in egocentric–wearable scenarios where the two modalities exhibit strong cross-modal spatio-temporal correlations, we argue that the strategy fails to operate effectively in exocentric–ambient environments due to two fundamental challenges.

\begin{figure}[tbp]  
    \centering
    \includegraphics[width=0.47\textwidth]{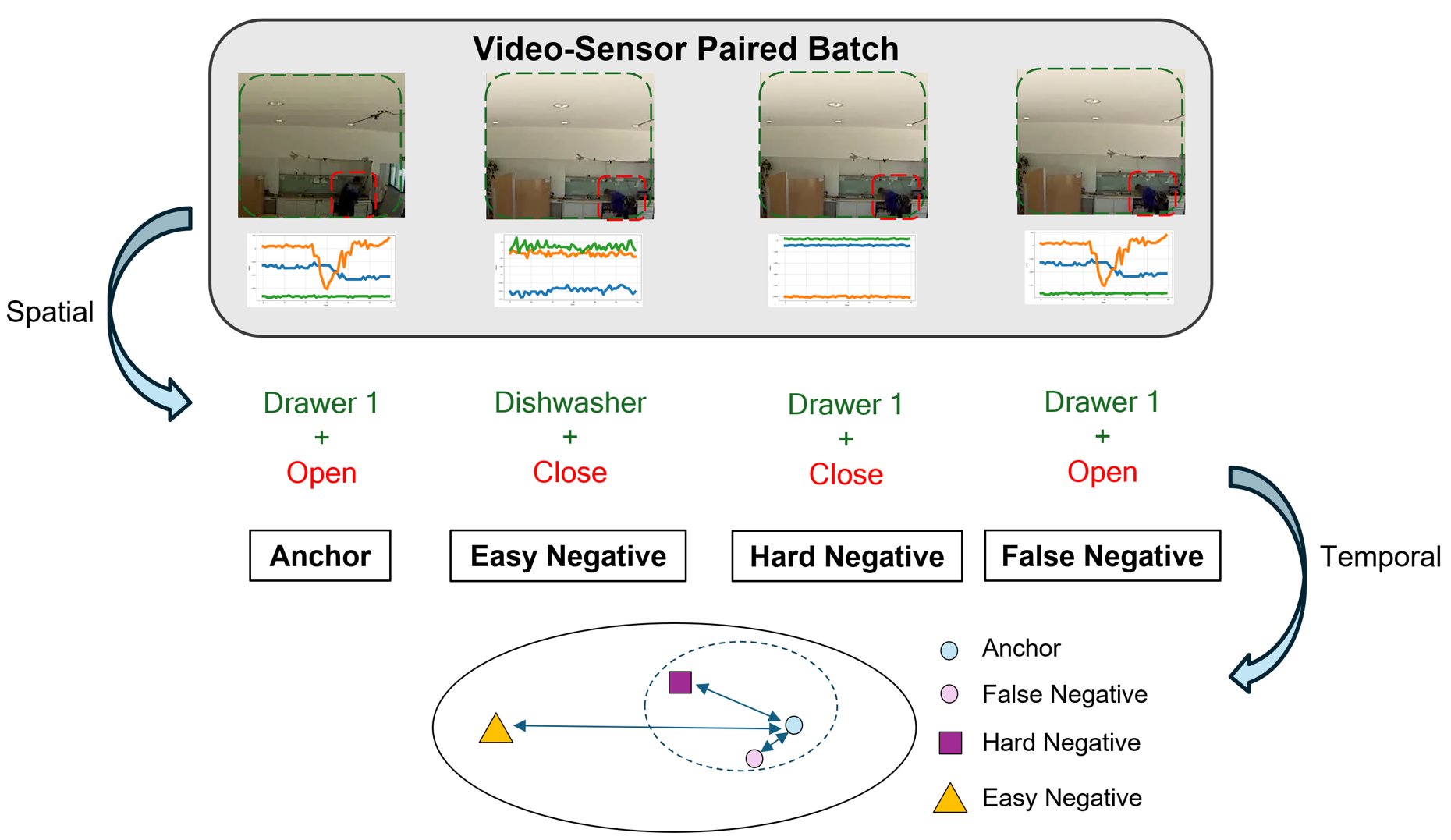}
    \vspace{-0.3cm}
    \caption{\textbf{Negative sampling strategy}. Hard negatives (same object, different actions) are prioritized, easy negatives (different objects, different actions) are down-weighted, and false negatives (same object, same action) are filtered.}
    \label{fig:NegativeMining}
\end{figure}

First, global alignment fails to capture fine-grained local details \cite{jin2023text}, specifically subtle motion patterns, in exocentric-ambient settings. 
Unlike egocentric video with salient hand-object interactions \cite{zhang2024masked}, exocentric video often shows minimal visual changes (e.g., a drawer opening slightly) against static scenes \cite{ardeshir2018exocentric, yin2024survey}. 
Compressing the entire video into a single vector dilutes these subtle yet discriminative cues, despite clear ambient sensor signals (Fig. ~\hyperref[fig:globalAlignment]{\ref*{fig:globalAlignment}(a)}).

Second, 
global alignment with motion-centric sensor encoders forces the model to rely solely on temporal patterns, which are the only modality-invariant feature \cite{jin2023text}. This creates a critical representational mismatch where video embeddings capture rich spatio-semantic cues (object appearance, spatial relationships, location), while sensor embeddings capture only temporal dynamics. Motion-centric encoders suit egocentric-wearable setups focused on body movements \cite{zhang2022deep, zhou2025efficient}, but exocentric-ambient settings require contextual grounding since each sensor is bound to a specific object or location. This leads to incorrect alignment across different spatio-semantic contexts (Fig. \hyperref[fig:globalAlignment]{\ref*{fig:globalAlignment}(b)}). 

To overcome these challenges, we propose DETACH, a framework for decomposed spatio-temporal alignment for exocentric video and ambient sensors with staged learning.
DETACH decomposes both modalities into spatial and temporal components. For video, it separates spatial context (object appearance, spatial relationships, location) from temporal motion, preserving subtle cues lost in Global Alignment. For sensors, it seperates temporal dynamics from spatial features that provide contextual grounding (e.g., the specific object or location). These sensor-spatial features are discovered via online clustering of activation patterns, enabling sensors to match the rich spatio-semantic context of video.

Reflecting the context-dependent nature of actions (e.g., open door), we introduce a two-stage, spatially-conditioned alignment. Stage 1 structures spatial features within each modality through cross-modal supervision, organizing representations by their spatio-semantic source. Stage 2 freezes these spatial features and aligns temporal features, forcing fine-grained temporal distinctions. We design a spatial-temporal weighted contrastive loss that prioritizes hard negatives while down-weighting easy negatives and filtering false negatives (Fig. \ref{fig:NegativeMining}). Our approach preserves fine-grained temporal cues and establishes semantic correspondence between modalities, directly resolving both limitations of global alignment in exocentric-ambient settings. 
Our main contributions are summarized as follows:
\begin{itemize}
    \item We identify two key limitations of Global Alignment in exocentric-ambient settings and propose a spatial-temporal decomposition framework (DETACH) to address them.
    \item We introduce a two-stage, spatially-conditioned alignment strategy that first establishes spatial grounding before aligning fine-grained temporal dynamics.
    \item We design a spatial-temporal weighted contrastive loss that leverages this spatial grounding to achieve fine-grained discrimination by prioritizing hard negatives while preventing false negatives.
    \item
    We demonstrate state-of-the-art performance on the Opportunity++ \cite{ciliberto2021opportunity++} and HWU-USP \cite{ranieri2021activity} datasets, achieving gains of up to 30\% in F1 and 43\% in mAP over existing egocentric–wearable baselines.
\end{itemize}
    \section{Related Work}
\label{sec:related_work}

\subsection{Human Action Recognition}
Our work falls within the broader field of Human Action Recognition (HAR), which can be categorized into vision-based and sensor-based approaches.

\noindent\textbf{Vision-based Human Action Recognition.}
Vision-based HAR can be categorized into egocentric (first-person) and exocentric (third-person) perspectives.3rd
Egocentric HAR captures fine-grained hand-object interactions from the user's viewpoint \cite{li2025challenges}, with research focusing on hand-object contact \cite{shiota2024egocentric, kapidis2019multitask} and gaze-based attention \cite{min2021integrating, huang2020mutual}. However, head-mounted cameras cause frequent target loss and face wearability constraints that limit practical use \cite{li2025challenges, tadesse2018visual}.
In contrast, exocentric HAR provides broader scene context and spatial relationships \cite{kong2022human}. The field has evolved from spatial-temporal models like two-stream architectures \cite{wang2016temporal, feichtenhofer2019slowfast} and 3D convolutions \cite{tran2015learning, carreira2017quo} to Transformer-based approaches \cite{arnab2021vivit, bertasius2021space}, commonly evaluated on Kinetics dataset \cite{kay2017kinetics}. However, the distant viewpoint limits fine-grained interaction capture and is sensitive to occlusion and lighting variations \cite{kong2022human}.

\noindent\textbf{Sensor-based Human Action Recognition.}
Sensor-based HAR has evolved along two parallel streams: body-centric wearable sensors and environment-centric ambient sensors.

Wearable sensor-based HAR primarily uses body-worn IMUs (accelerometers and gyroscopes) to capture fine-grained movements. Methods have advanced from hand-crafted features \cite{bao2004activity} through CNNs \cite{jiang2015human}, RNNs \cite{murad2017deep}, and hybrid models \cite{ordonez2016deep, yao2017deepsense} to Transformer architectures \cite{mahmud2020human}. While wearable sensors excel at measuring precise bodily kinematics, they lack environmental context such as object interactions and state changes \cite{ni2024survey}.
In contrast, ambient sensor-based HAR uses diverse sensors to detect object or space state changes \cite{van2008accurate, rashidi2012survey}, including event-based devices (contact/pressure sensors) \cite{tapia2004activity, van2008accurate}, state-based devices (power consumption) \cite{biansoongnern2016non}, and object-attached IMUs \cite{nabiei2015object}. However, many ambient sensors provide only coarse or discrete signals (e.g., on/off), leaving underlying actions ambiguous and limiting full behavior understanding.

\noindent\textbf{Egocentric Multimodal Human Action Recognition.}
Recent studies integrate wearable sensors with egocentric video to leverage complementary strengths, as video provides hand-object interactions, while sensors capture subtle motion dynamics. IMU2CLIP \cite{moon2022imu2clip} extended CLIP \cite{radford2021learning} to IMUs through contrastive learning. PRIMUS \cite{das2025primus} improved this by using video-based nearest neighbors as positives. COMODO \cite{chen2025comodo} introduced cross-modal distillation, where a frozen video encoder guides the IMU encoder. EVI-MAE \cite{zhang2024masked}  combined contrastive learning with masked autoencoding to jointly reconstruct both modalities.

These methods commonly employ Global Alignment, encoding entire video and sensor sequences into single vectors to align them. This approach proves effective in egocentric-wearable settings where user motion inherently synchronizes both modalities \cite{zhang2024masked}. However, alignment between exocentric videos and ambient sensors remains unexplored, presenting new challenges rooted in the fundamental limitations of global alignment.

\subsection{Cross-modal Alignment Paradigms}
While Global Alignment prevails in cross-modal fields like Text-Video Retrieval \cite{gabeur2020multi, bain2021frozen, luo2022clip4clip}, it inherently lacks the capacity to capture local details and tends to over-rely on modality-invariant features \cite{jin2023text}. To address these limitations, recent studies have shifted towards fine-grained alignment via semantic decomposition \cite{jin2023text, li2023progressive}, enabling precise correspondence learning. We extend this granular paradigm to the exocentric-ambient setting, proposing to decompose and align temporal and spatial cues from both modalities.
     \section{Method}
\label{sec:method}
\begin{figure*}[htbp]
    \centering
    \includegraphics[width=\textwidth]{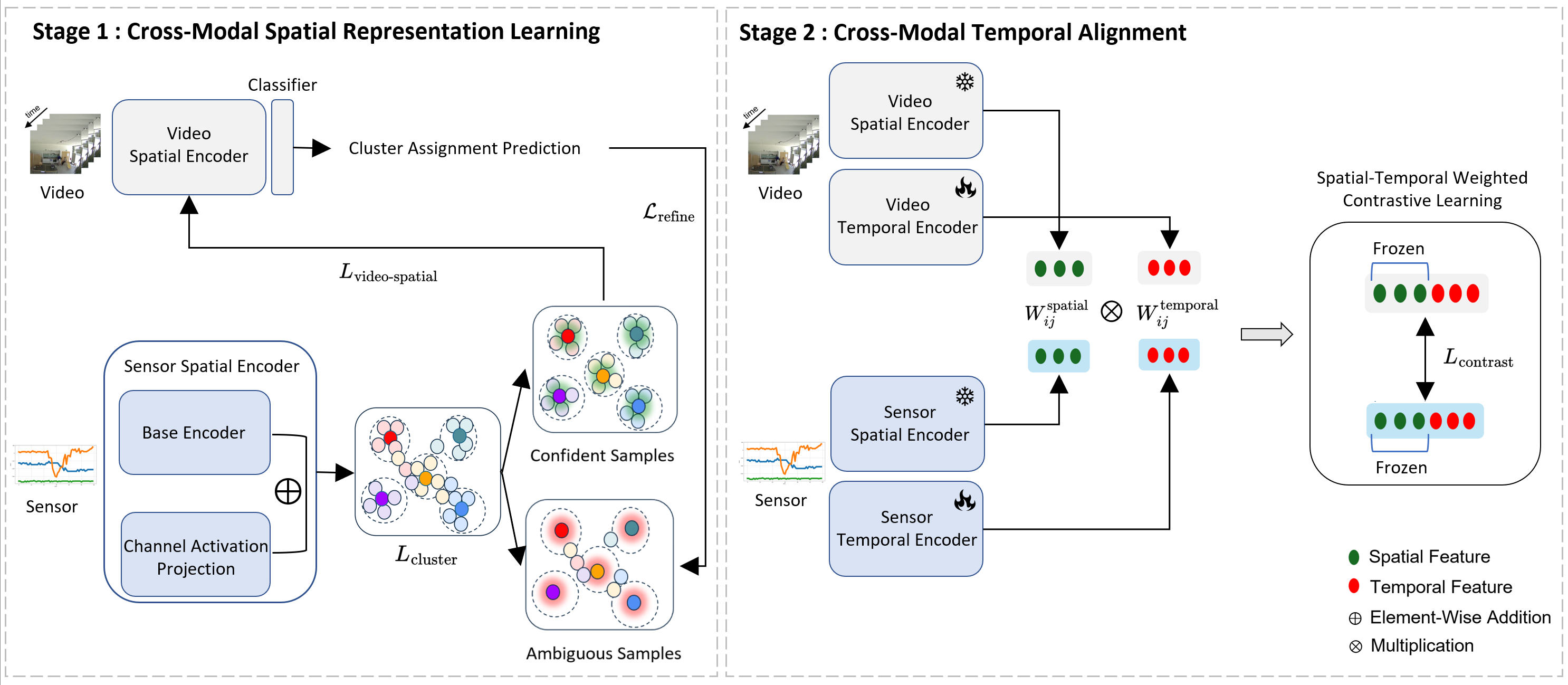}
    \vspace{-0.7cm}
    \caption{\textbf{Overall Architecture of DETACH}. Stage 1 decomposes and learns spatial representations. Stage 2 leverages these spatial features to guide temporal alignment by decomposing temporal dynamics. This strategy enables fine-grained cross-modal understanding by separately capturing spatial context and temporal patterns.}
    \label{fig:Architecture}
\end{figure*}

Our framework employs a two-stage decomposition approach for cross-modal alignment (Fig. \ref{fig:Architecture}). Stage 1 learns spatial representations, and Stage 2 aligns temporal dynamics guided by the learned spatial features.

\subsection{Stage 1: Cross-Modal Spatial Representation Learning}
Stage 1 focuses on learning spatially-structured representations for both modalities. We employ a three-phase approach: (1) online sensor clustering to discover spatial activation patterns, (2) joint learning where video spatial features are trained on high-confidence sensor assignments, and (3) video-guided refinement to resolve ambiguous clusters.

\noindent\textbf{Spatial Feature Extraction via Sensor Clustering.}
Ambient sensors are spatially distributed, with each activation indicating activity at a specific location. When a sensor is triggered, its channel shows significant change while others remain inactive \cite{krishnan2014activity}, creating location-specific activation patterns that we treat as spatial features. We adopt online clustering \cite{zhan2020online} to learn these spatial features, with cluster count matching the number of sensor sources for one-to-one correspondence between clusters and sensor locations.

We first encode sensor through a base encoder combining 1D CNN and GRU to capture activation patterns. To enhance spatial distinctiveness, we incorporate channel activation intensity as a complementary cue by computing the 99th percentile of signal values for each channel, which reduces outlier sensitivity. This statistical vector is then projected and fused with the encoder output via element-wise addition. Following ODC \cite{zhan2020online}, we jointly optimize the sensor spatial encoder and learnable cluster centroids using a memory bank maintaining consistent pseudo-labels. The clustering objective is defined as a class-balanced cross-entropy:

\begin{equation}
\mathcal{L}_{\text{cluster}} = -\frac{1}{N}\sum_{i=1}^{N} w_{y_i} \log(p_{i, y_i})
\end{equation}

where $N$ is the batch size, $y_i$ is the pseudo-label, $p_{i, y_i}$ is the softmax probability for cluster $y_i$, and \(w_{y_i} = 1 / |C_{y_i}|^{0.5}\) is the inverse-frequency weight for cluster \(y_i\) to address cluster imbalance. This loss jointly optimizes the encoder and the learnable cluster centroids.

\noindent\textbf{Video Spatial Feature Learning with Confident Pseudo-Labels.}
To enable video representations to learn spatial patterns corresponding to sensor clusters, we extract frame-level spatial features. We temporally average video frames, process it through a 2D convolutional encoder, and feed the extracted features into a linear classifier predicting sensor-derived pseudo-labels.

However, ambient sensors may produce false activations \cite{wilhelm2024emergency} and unsupervised clustering may yield noisy assignments \cite{van2020scan}, making not all samples reliable for training. Inspired by prototype-based clustering \cite{snell2017prototypical}, we identify high-quality samples using Euclidean distance to cluster centroids by marking samples below the 75th percentile distance as confident and others as ambiguous. This per-cluster criterion accounts for cluster density variations.

During joint learning ($T$ epochs), sensor clustering continues on all samples while the video spatial encoder trains only on confident samples. Each confident sample's video is processed to predict its pseudo-label via cross-entropy:

\begin{equation}
\mathcal{L}_{\text{video-spatial}} = - \frac{1}{N_{\text{conf}}} \sum_{i \in \text{conf}} \log(p_{i, y_i})
\end{equation}

where ${N_{\text{conf}}}$ is the number of confident samples and ${y_i}$ is the pseudo-label from sensor clustering.

\noindent\textbf{Video-Guided Refinement.}
After joint learning, the video spatial encoder has acquired reliable spatial representations. We freeze the encoder and use its predictions to refine sensor clustering on ambiguous samples, guiding sensor cluster assignments toward higher accuracy. The refinement loss applies cross-entropy between the sensor predictions and video-derived hard pseudo-labels on ambiguous samples:

\begin{equation}
\mathcal{L}_{\text{refine}} = - \frac{1}{N_{\text{amb}}} \sum_{i \in \text{amb}} \log(p_{i, \hat{y}_i}^{\text{sensor}}), \quad
\hat{y}_i = \arg \max_{k} p_{i,k}^{\text{video}}
\end{equation}

where $N_{\text{amb}}$ is the number of ambiguous samples, $p_{i, \hat{y}_i}^{\text{sensor}}$ is the sensor spatial encoder's predicted probability for cluster $\hat{y}_i$, and $\hat{y}_i$ is the hard pseudo-label from the frozen video spatial encoder. 

The total sensor loss combines clustering (on all samples) and refinement (on ambiguous samples) objectives:

\begin{equation}
\mathcal{L}_{\text{sensor}} = \alpha \mathcal{L}_{\text{cluster}} + \beta \mathcal{L}_{\text{refine}}
\label{eq:stage1_sensor_loss}
\end{equation}

where $\alpha$ and $\beta$ balances the two objectives. 


\subsection{Stage 2: Cross-Modal Temporal Alignment}
Stage 2 focuses on learning temporal representations that distinguish fine-grained actions within the established spatial context. We employ a three-phase approach: (1) spatial-conditioned temporal encoding combining frozen spatial features with trainable temporal encoders, (2) adaptive weighting that up-weights hard negatives while down-weighting false negatives, and (3) spatial-temporal weighted contrastive learning.

\noindent\textbf{Spatially-Conditioned Temporal Alignment.}
Building on Stage 1's spatial representations, we focus on learning temporal features. Temporal alignment must be spatially-conditioned, as distinguishing ``open door" from ``close door" requires the door context. We combine frozen spatial features with trainable temporal features for each modality:

\begin{align}
\mathbf{z}_{\text{video}} &= [\mathbf{v}_{\text{spatial}}, \mathbf{v}_{\text{temporal}}] \in \mathbb{R}^{2d}, \\
\mathbf{z}_{\text{sensor}} &= [\mathbf{s}_{\text{spatial}}, \mathbf{s}_{\text{temporal}}] \in \mathbb{R}^{2d}
\end{align}
where spatial and temporal features are both $d$-dimensional.

\noindent\textbf{Spatial-Aware Hard Negative Mining.}
We apply contrastive learning to align these spatially-conditioned representations. However, standard contrastive learning treats all negatives equally \cite{oord2018representation}, causing models to rely on easy distinctions \cite{chuang2020debiased} between different spatial contexts rather than subtle temporal differences. We address this through spatial-aware hard negative mining with adaptive weighting \cite{robinson2021contrastive}. Easy negatives with dissimilar spatial features are already distinguishable through frozen spatial encoders. Hard negatives with similar spatial features but different temporal dynamics require fine-grained motion discrimination. Our adaptive weighting emphasizes hard negatives, forcing temporal encoders to focus on motion patterns within the same spatial context \cite{robinson2021contrastive}.

Since video and sensor spatial features reside in separate latent spaces, we define the combined spatial similarity using a max operation to avoid scale inconsistencies from averaging. We apply ReLU to exclude dissimilar pairs with negative cosine similarities, as these are easily distinguishable and should not contribute to hard negative weighting. A pair is considered spatially similar if either modality indicates high similarity. For any sample pair $(i,j)$, this is computed as: 

\begin{equation} 
s_{ij}^{\text{spatial}} = \max(\text{ReLU}(s_{ij}^{\text{v-spatial}}), \text{ReLU}(s_{ij}^{\text{s-spatial}})) 
\end{equation}

where $s_{ij}^{\text{v-spatial}}$ and $s_{ij}^{\text{s-spatial}}$ are the cosine similarities of normalized spatial features for video and sensor modalities, respectively. 

Based on spatial similarity, we assign adaptive weights to negative pairs:

\begin{equation}
W_{ij}^{\text{spatial}} = 1.0 + (\lambda_{\text{hard}} - 1.0) \cdot s_{ij}^{\text{spatial}}, \quad \lambda_{\text{hard}} > 1
\end{equation} 

where $\lambda_{\text{hard}}$ controls hard negative importance. This weight interpolates from 1.0 (easy negatives, $s_{ij}^{\text{spatial}} \approx 0$) to $\lambda_{\text{hard}}$ (hard negatives, $s_{ij}^{\text{spatial}} \approx 1$).

\noindent\textbf{Temporal-Aware False Negative Mitigation.}
However, weighting solely on spatial similarity introduces false negatives, which are pairs sharing both spatial and temporal characteristics that should not be strongly penalized \cite{chuang2020debiased, villegas2017decomposing}. We mitigate this by reducing weights for temporally similar pairs. Temporal features are extracted by modality-specific encoders. For video, we use 3D convolutions on frame differences and optical flow following video decomposition approaches \cite{sun2023moso, villegas2017decomposing}. For sensor data, we adapt the 1D CNN-GRU-attention architecture based on prior works \cite{moon2022imu2clip, das2025primus}. Computing temporal similarity using online temporal encoders introduces instability, as representations continuously change \cite{he2020momentum, grill2020bootstrap}. We address this with momentum encoders with EMA updates ($m$=0.999) for stable estimates. Since both modalities' temporal features are aligned in a shared space, combined temporal similarity can be defined by directly averaging them:

\begin{equation}
s_{ij}^{\text{temporal}} = (s_{ij}^{\text{v-temporal}} + s_{ij}^{\text{s-temporal}}) / 2
\end{equation}

where $s_{ij}^{\text{v-temporal}}$ and $s_{ij}^{\text{s-temporal}}$ denote cosine similarities for video and sensor temporal features.

The temporal similarity-based weight is then defined as:

\begin{equation} 
W_{ij}^{\text{temporal}} = 1.0 - s_{ij}^{\text{spatial}} \cdot \text{ReLU}(s_{ij}^{\text{temporal}}) 
\end{equation}

We apply ReLU for two objectives, proportionally reducing weights based on temporal similarity degree and excluding pairs with dissimilar motion patterns where cosine similarity is negative. This ensures pairs with high spatial and temporal similarity receive reduced penalties, mitigating false negative over-penalization.

The final spatial-temporal adaptive weight combines both components:

\begin{equation}
W_{ij} = W_{ij}^{\text{spatial}} \cdot W_{ij}^{\text{temporal}}, \quad \text{with} \quad W_{ii} = 0
\end{equation}

\noindent\textbf{Spatial-Temporal Weighted Contrastive Learning.}
We integrate the spatial-temporal adaptive weights into a bidirectional InfoNCE loss \cite{radford2021learning}, where each video representation is contrasted against all sensor representations, and vice versa. The weighted contrastive loss for the video-to-sensor direction is defined as:

\begin{align}
s_{ij} &= (\mathbf{z}_{\text{video}}^{(i)} \cdot \mathbf{z}_{\text{sensor}}^{(j)}) / \tau_{\text{contrast}} \nonumber \\
\mathcal{L}_{\text{v2s}} &= -\frac{1}{N}\sum_{i=1}^{N} \log \frac{\exp(s_{ii})}{\exp(s_{ii}) + \sum_{j \neq i} W_{ij} \cdot \exp(s_{ij})}
\end{align}

where $W_{ij}$ are the spatial-temporal adaptive weights. 

The sensor-to-video loss $\mathcal{L}_{\text{s2v}}$ is computed symmetrically, and the final bidirectional contrastive loss is their average:

\begin{equation}
\mathcal{L}_{\text{contrast}} = \frac{1}{2}(\mathcal{L}_{\text{v2s}} + \mathcal{L}_{\text{s2v}})
\end{equation}
    \section{Experiments}

\noindent\textbf{Datasets.}
We conduct experiments on two datasets: Opportunity++ \cite{ciliberto2021opportunity++} and HWU-USP \cite{ranieri2021activity}. Both datasets capture

\begin{table*}[t!]
    \centering
    \footnotesize 
    \renewcommand{\arraystretch}{0.9} 
    \caption{\textbf{Downstream classification performance comparison with baseline methods.} 
    Best results are in \textbf{bold}. Values in {\tiny\textcolor{red}{($\downarrow$)}}  denote the performance gap from our method.}
    \label{tab:main_performance_delta}
    \vspace{-0.3cm}
    
    \begin{tabular*}{\textwidth}{l @{\extracolsep{\fill}} cccc} 
        \toprule
        \multirow{2}{*}{\textbf{Model}} & \multicolumn{2}{c}{\textbf{Opportunity++} \cite{ciliberto2021opportunity++}} & \multicolumn{2}{c}{\textbf{HWU-USP} \cite{ranieri2021activity}} \\
        \cmidrule(lr){2-3} \cmidrule(lr){4-5}
         & F1 (Weighted) & mAP & F1 (Weighted) & mAP \\ 
        \midrule
        
        IMU2CLIP \cite{moon2022imu2clip} 
        & 0.39 \, \tiny\textcolor{red}{($\downarrow$ 0.34)} 
        & 0.43 \, \tiny\textcolor{red}{($\downarrow$ 0.40)} 
        & 0.50 \, \tiny\textcolor{red}{($\downarrow$ 0.21)} 
        & 0.64 \, \tiny\textcolor{red}{($\downarrow$ 0.03)} \\
        
        PRIMUS \cite{das2025primus}
        & 0.28 \, \tiny\textcolor{red}{($\downarrow$ 0.45)}
        & 0.31 \, \tiny\textcolor{red}{($\downarrow$ 0.52)}
        & 0.48 \, \tiny\textcolor{red}{($\downarrow$ 0.23)}
        & 0.61 \, \tiny\textcolor{red}{($\downarrow$ 0.06)} \\

        COMODO \cite{chen2025comodo}
        & 0.43 \, \tiny\textcolor{red}{($\downarrow$ 0.30)}
        & 0.52 \, \tiny\textcolor{red}{($\downarrow$ 0.31)}
        & 0.50 \, \tiny\textcolor{red}{($\downarrow$ 0.21)}
        & 0.64 \, \tiny\textcolor{red}{($\downarrow$ 0.03)} \\

        CAV-MAE \cite{gongcontrastive} 
        & 0.56 \, \tiny\textcolor{red}{($\downarrow$ 0.17)}
        & 0.58 \, \tiny\textcolor{red}{($\downarrow$ 0.25)}
        & 0.70 \, \tiny\textcolor{red}{($\downarrow$ 0.01)}
        & 0.55 \, \tiny\textcolor{red}{($\downarrow$ 0.12)} \\
        \midrule
        
        \bfseries DETACH (Ours) 
        & \bfseries 0.73 & \bfseries 0.83 & \bfseries 0.71 & \bfseries 0.67 \\
        \bottomrule
    \end{tabular*}
\end{table*}

\begin{table*}[t!]
    \centering
    
    \begin{minipage}[t]{0.48\textwidth}
        \centering
        \caption{\textbf{Ablation on spatial-temporal adaptive weight components.} We evaluate the impact of removing 
        \label{tab:ablation_weights}$W_{ij}^{\text{spatial}}$ and $W_{ij}^{\text{temporal}}$ from the full weighting scheme.}
        \label{tab:ablation_weights}

        \vspace{-0.3cm}
        
        \scalebox{0.80}{
            \begin{tabular}{lcccc}
                \toprule
                \multirow{2}{*}{\textbf{Method}} & \multicolumn{2}{c}{\textbf{Opportunity++}} & \multicolumn{2}{c}{\textbf{HWU-USP}} \\
                \cmidrule(lr){2-3} \cmidrule(lr){4-5}
                 & F1 (Weighted) & mAP & F1 (Weighted) & mAP \\
                \midrule
                $L_{contrast}$ & \textbf{0.73} & \textbf{0.83} & \textbf{0.71} & \textbf{0.67} \\
                w/o $W_{ij}^{\text{spatial}}$ & 0.56 & 0.71 & 0.65 & 0.61 \\
                w/o $W_{ij}^{\text{temporal}}$ & 0.62 & 0.71 & 0.60 & 0.61 \\
                \bottomrule
            \end{tabular}
        }
    \end{minipage}
    \hfill
    \begin{minipage}[t]{0.48\textwidth}
        \centering
        \caption{{\textbf{Ablation on loss components.} We analyze the impact of the refinement loss ($L_{refine}$) and our sptial-temporal weighted contrastive loss ($L_{contrast}$) against simple InfoNCE.}}
        \label{tab:ablation_loss}

        \vspace{-0.3cm}
        
        \scalebox{0.72}{ 
            \begin{tabular}{clcccc}
                \toprule
                \multirow{2}{*}{\textbf{Refinement}} & \multirow{2}{*}{\textbf{Contrastive}} & \multicolumn{2}{c}{\textbf{Opportunity++}} & \multicolumn{2}{c}{\textbf{HWU-USP}} \\
                \cmidrule(lr){3-4} \cmidrule(lr){5-6}
                 & & F1 (Weighted) & mAP & F1 (Weighted) & mAP \\
                \midrule
                \checkmark & $L_{contrast}$ & \textbf{0.73} & \textbf{0.83} & \textbf{0.71} & \textbf{0.67} \\
                \checkmark & InfoNCE & 0.59 & 0.73 & 0.49 & 0.62 \\
                \texttimes & $L_{contrast}$ & 0.60 & 0.70 & 0.69 & 0.65 \\
                \texttimes & InfoNCE & 0.59 & 0.70 & 0.68 & 0.62 \\
                \bottomrule
            \end{tabular}
        }
    \end{minipage}

    \vspace{0.3cm} 
    
\end{table*}

\noindent participants interacting with various objects in indoor environments with fixed backgrounds. \textbf{1) Opportunity++}  includes IMUs and logic switches attached to objects. Videos are recorded at 10 fps, while sensors operate at 30 Hz. We selected 14 mid-level labels that utilize ambient sensors, structured as open/close actions for 7 objects. \textbf{2) HWU-USP} includes logic switches and logic motion sensors attached to objects or walls. Videos are recorded at 25 fps, while sensors operate at 50 Hz. This dataset only provides high-level labels  for entire 40-50 seconds sequences, without fine-grained temporal annotations. We selected 5 labels where ambient sensors provide discriminative signals for action recognition.

\noindent\textbf{Preprocessing.} We apply a sliding window of 2 seconds with 1 second overlap, following related works \cite{miao2024goat, haresamudram2024towards}. For Opportunity++, the resulting 5K samples are split into pretraining (80\%) and downstream evaluation (20\%), with the latter further divided into train/val/test (80\%/10\%/10\%). For HWU-USP, to prevent data leakage, we first split the raw files into pretraining (70\%) and downstream evaluation (30\%) considering label distribution. Then windowing is applied to these files, resulting in a total of 3K samples. The samples for downstream evaluation are then divided into train/test (80\%/20\%). Sensor data is z-score normalized per channel.

\begin{table*}[t!]
    \begin{minipage}[t]{0.48\textwidth}
        \centering
        \caption{\textbf{Ablation on the momentum encoder used for calculating $W_{ij}^{\text{temporal}}$.}}
        \label{tab:ablation_momentum}

        \vspace{-0.3cm}
        
        \scalebox{0.7}{
            \begin{tabular}{lcccc}
                \toprule
                \multirow{2}{*}{\textbf{Method}} & \multicolumn{2}{c}{\textbf{Opportunity++}} & \multicolumn{2}{c}{\textbf{HWU-USP}} \\
                \cmidrule(lr){2-3} \cmidrule(lr){4-5}
                 & F1 (Weighted) & mAP & F1 (Weighted) & mAP \\
                \midrule
                w Momentum Encoder (Ours) & \textbf{0.73} & \textbf{0.83} & \textbf{0.71} & \textbf{0.67} \\
                w/o Momentum Encoder & 0.59 & 0.72 & 0.65 & 0.61 \\
                \bottomrule
            \end{tabular}
        }
    \end{minipage}
    \hfill
    \begin{minipage}[t]{0.48\textwidth}
        \centering
        \caption{\textbf{Ablation on the confidence threshold for selecting confident samples in the clustering stage.}}
        \label{tab:ablation_threshold}

        \vspace{-0.3cm}
        
        \scalebox{0.75}{
            \begin{tabular}{lcccc}
                \toprule
                \multirow{2}{*}{\textbf{Threshold}} & \multicolumn{2}{c}{\textbf{Opportunity++}} & \multicolumn{2}{c}{\textbf{HWU-USP}} \\
                \cmidrule(lr){2-3} \cmidrule(lr){4-5}
                 & F1 (Weighted) & mAP & F1 (Weighted) & mAP \\
                \midrule
                75\% (Ours) & \textbf{0.73} & \textbf{0.83} & \textbf{0.71} & \textbf{0.67} \\
                100\% & 0.53 & 0.71 & 0.60 & 0.63 \\
                \bottomrule
            \end{tabular}
        }
    \end{minipage}
    
\end{table*}

\noindent\textbf{Baselines.}
We compare our framework with state-of-the-art multimodal methods, IMU2CLIP \cite{moon2022imu2clip}, PRIMUS \cite{das2025primus}, COMODO \cite{chen2025comodo} and EVI-MAE \cite{zhang2024masked}, adapted for the exocentric-ambient setting. For IMU2CLIP and PRIMUS, we exclude text branch to ensure fair comparison based solely on video and sensor modalities. For EVI-MAE, we use the corresponding backbone CAV-MAE \cite{gongcontrastive} as a baseline, since EVI-MAE's graph structure for wearable sensors is inapplicable to ambient sensors with high-dimensional nodes.

\noindent\textbf{Implementation Details.}
For training, we use the AdamW optimizer (weight decay $1 \times 10^{-4}$, initial learning rate $1 \times 10^{-4}$) and a batch size of 256. We set $K=7$ clusters for both datasets, where HWU-USP represents 6 object interactions and non-interactive state. For sensor loss, the balancing weights are $\alpha=1.0$ and $\beta=1.5$. The pretraining stage runs for 50 epochs, and the joint learning phase runs for $T=10$ epochs. Contrastive learning uses $\tau_{\text{contrast}} = 0.10$ and $\lambda_{hard} = 3.0$. 

\noindent\textbf{Evaluation Protocol.}
We evaluate the quality of sensor representations through downstream classification task. After pretraining for Opportunity++, we freeze the sensor encoder and attach a linear classifier for 14-class mid-level action classification, as each 2-second window has a corresponding label. For HWU-USP, we freeze the pretrained sensor encoder and extract features from 2-second windows, which are temporally pooled across the sequence as in \cite{wang2016temporal}, and fed into a classifier for 5-class high-level action classification. 

\noindent\textbf{Evaluation Metrics.}
We report weighted F1-score and mean Average Precision (mAP) as evaluation metrics to account for class imbalance.

\subsection{Main Results}
Tab. {\ref{tab:main_performance_delta}} confirms that DETACH achieves state-of-the-art performance across all metrics. On Opportunity++ \cite{ciliberto2021opportunity++}, it outperforms the second-best model CAV-MAE \cite{gongcontrastive}, by +30.4\% in weighted F1 and +43.1\% in mAP. On HWU-USP \cite{ranieri2021activity}, our method also yields +1.4\% improvement in weighted F1 and +21.8\% in mAP over CAV-MAE \cite{gongcontrastive}. While CAV-MAE \cite{gongcontrastive} benefits from multimodal masked modeling, its reliance on Global Alignment remains a  

\begin{figure*}[htbp]
    \centering
    \includegraphics[width=\textwidth]{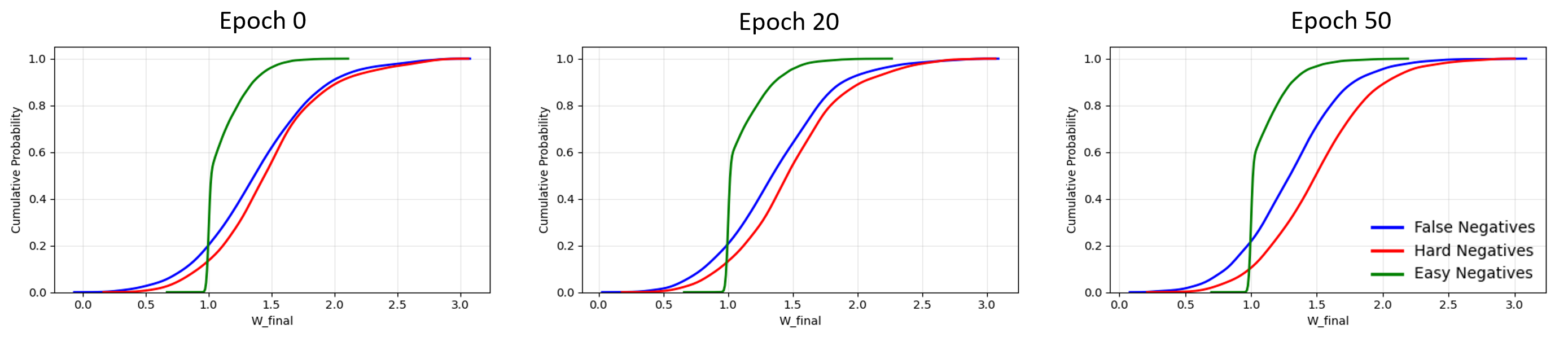}
    \vspace{-0.7cm}
    \caption{\textbf{Temporal evolution of negative sample discrimination}. This figure represents cumulative distribution functions (CDFs) of final weights $W_{ij}$ across training epochs. Initially overlapped distributions of hard and false negatives gradually separate, while easy negatives maintain unit weights throughout training.}
    \label{fig:cdf}
\end{figure*}

\noindent critical bottleneck. In contrast, DETACH effectively resolves the limitations of Global Alignment through spatial-temporal decomposed alignment, enabling fine-grained understanding for complex exocentric-ambient settings.

\subsection{Ablation Studies}

\noindent\textbf{Effectiveness of Spatial-Temporal Adaptive Weight Components.} 
Tab. \ref{tab:ablation_weights} analyzes the contribution of each component for spatial-temporal adaptive weighting. Removing $W_{ij}^{\text{spatial}}$, which is desinged for hard negative mining, leads to substantial performance drop on Opportunity++. This suggests that spatial context differentiation is critical for datasets with complex backgrounds. Similarly, excluding $W_{ij}^{\text{temporal}}$, which targets false negative mitiagion, consistently degrades performance across both benchmarks. These results validate that both components play complementary roles in robust cross-modal alignment.

\noindent\textbf{Impact of Loss Components.} 
Tab. \ref{tab:ablation_loss} demonstrates the contributions of both $\mathcal{L}_{refine}$ and  $\mathcal{L}_{contrast}$. Unlike standard InfoNCE treating all samples equally, $\mathcal{L}_{contrast}$ significantly outperforms the baseline by using spatial-temporal adaptive weights to prioritize hard negatives and mitigate false negatives. Meanwhile, $\mathcal{L}_{refine}$  enables mutual correction between modalities, with the video encoder refining the sensor encoder to compensate for uncertainties. These results confirm that both components are essential for robust alignment.

\noindent\textbf{Necessity of the Momentum Encoder.}
Tab. \ref{tab:ablation_momentum} examines the necessity of the momentum encoder for computing $W_{ij}^{\text{temporal}}$. Directly using the training temporal encoder causes significant performance drops due to unstable similarity estimations. The momentum encoder provides stable feature representations, enabling reliable temporal similarity measurement and effective false negative mitigation through precise $W_{ij}^{\text{temporal}}$ weighting.

\noindent\textbf{Importance of Confidence-based Filtering.}
Tab. \ref{tab:ablation_threshold} highlights the critical role of confidence thresholding. Utilizing all samples (100\%) causes drastic degradation on both datasets, indicating that indiscriminate usage of samples injects noisy supervision. Our strategy effectively filters these ambiguous predictions, ensuring that only high-confidence pseudo-labels drive the cross-modal alignment.

\subsection{Discussion and Analysis}

\noindent\textbf{Analysis on the Separation of Hard and False Negatives.}
We visualized the cumulative distribution functions (CDFs) of final weights $W_{ij}$ across epochs (Fig. \ref{fig:cdf}) to analyze the model's evolving discrimination capability. Initially, hard and false negatives show overlapped distributions, indicating poor differentiation. In constrast, easy negatives consistently maintain unit weights ($W_{ij} \approx 1.0$), demonstrating training stability. As training progresses, a clear statistical divergence emerges between the two groups, with false negatives shifting to a significantly lower weight range than hard negatives. While false negative weights do not reach near-zero values, this relative separation confirms our approach leverages temporal consistency to distinguish false negatives from hard negatives.

\begin{figure}[htbp]  
    \centering
    \includegraphics[width=0.47\textwidth]{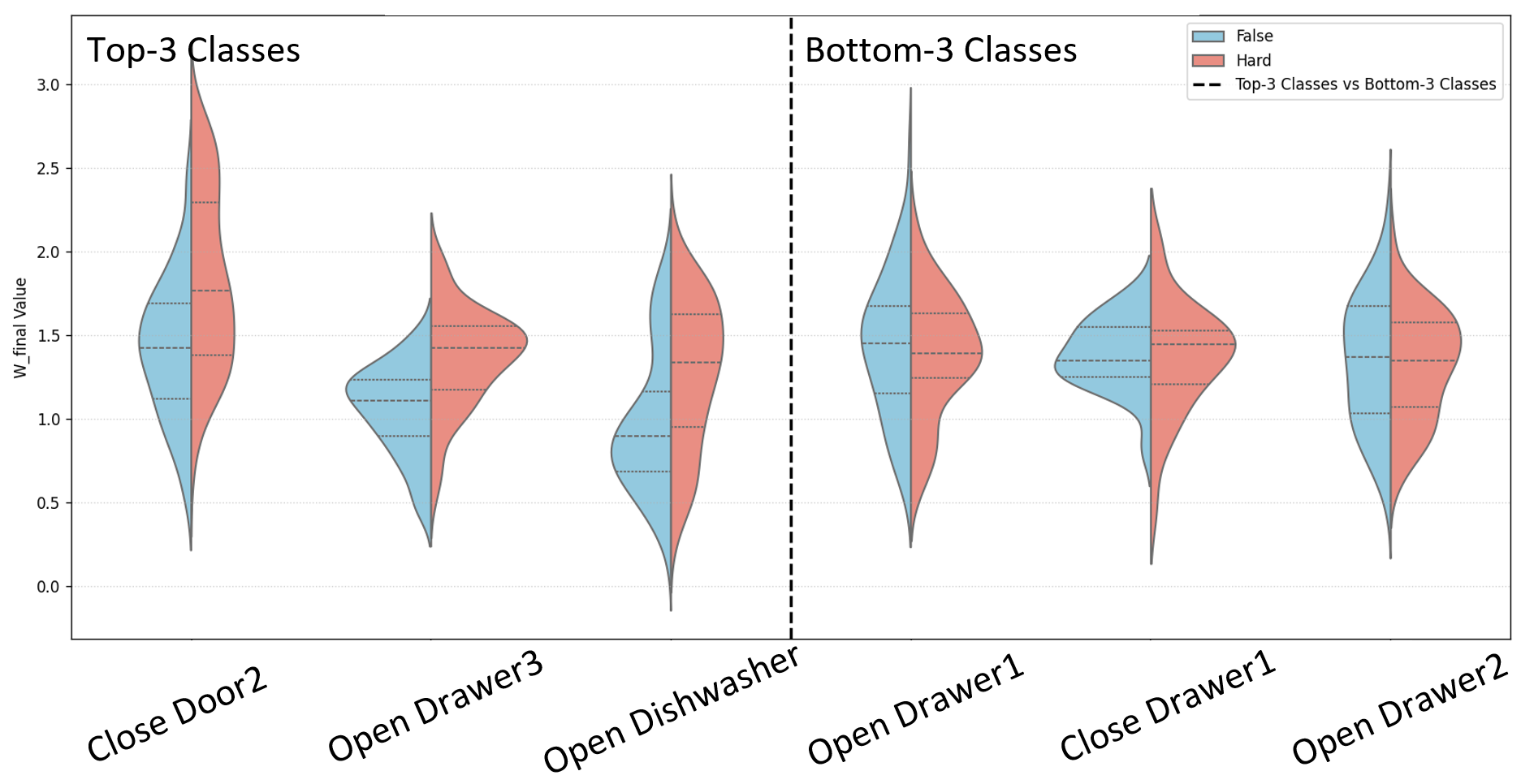}
    \vspace{-0.3cm}
    \caption{\textbf{Class-wise analysis of hard and false negative separation}. Comparison of weight distributions between top-3 classes with the largest separation (close door2, open drawer3, open dishwasher) and bottom-3 classes with the smallest separation (open drawer1, close drawer1, open drawer2).}
    \label{fig:violet}
\end{figure}

\noindent\textbf{Class-wise Analysis of Hard and False Negatives Separation.}
We further investigated how the relative separation varies across different action labels (Fig. \ref{fig:violet}). We compared the top-3 classes showing the largest weight differences against the bottom-3 classes with the smallest differences. Upon closer inspection, classes with significant separation tend to display clear and high-magnitude movements, while those with minimal separation appear to

\noindent involve subtle or nearly static dynamics. As these distinct motion types coexist within the dataset, the model struggles to strictly separate false negatives in static scenarios. This data heterogeneity prevents ideal convergence, resulting in the relative separation.

\begin{figure}[tbp]  
    \centering
    \includegraphics[width=0.47\textwidth]{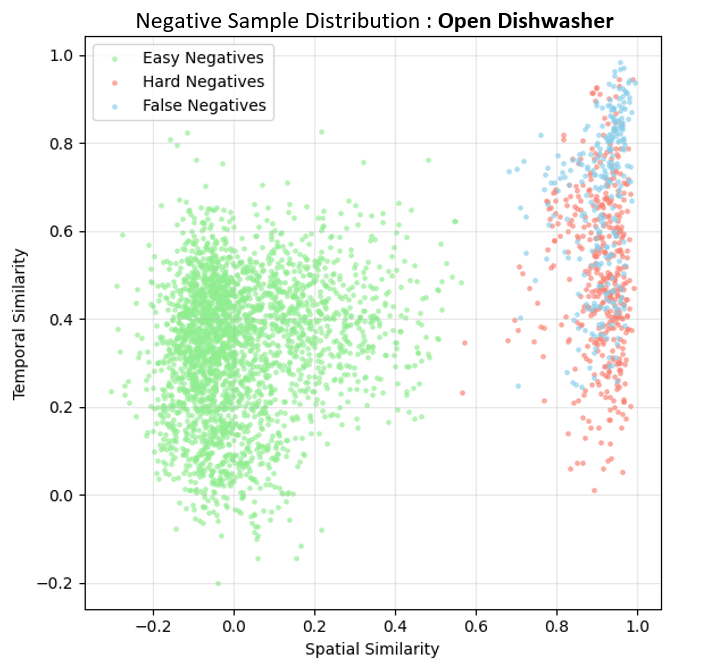}
    \vspace{-0.3cm}
    \caption{\textbf{Visualization of negative samples in feature space}. Feature relationship phase space for a query sample ``Open Dishwasher" with spatial similarity (x-axis) and temporal similarity (y-axis). Hard and false negatives are separated along the temporal axis despite high spatial similarity.}
    \label{fig:scatter}
\end{figure}

\noindent\textbf{Visualization of Negative Samples in Feature Space.}
To provide a concrete understanding of how our model handles hard negative samples, we visualize the feature relationship phase space for a representative query sample, ``Open Dishwasher" (Fig. \ref{fig:scatter}). The x and y-axes represent spatial and temporal similarities, respectively. Easy negatives form a distinct cluster in the low similarity region (left side). In contrast, hard negatives and false negatives are spatially clustered in the high-appearance similarity region (right side), yet clear separation emerges along the temporal axis. The model distinguishes false negatives (top-right) by their high temporal similarity from hard negatives (bottom-right) which exhibit low temporal similarity. This confirms our method effectively utilizes temporal cues to distinguish false negatives from hard negatives in challenging scenarios.

\noindent\textbf{Visualization of Video and Sensor Feature Spaces.}
We visualize the feature spaces of both modalities using t-SNE (Fig. \ref{fig:tsne}). The visualizations consistently reveal a context-dependent structure across both modalities. At a global level, samples are primarily grouped by spatial context. Within these spatial clusters, fine-grained differentiation emerges based on temporal context, where distinct actions form well-separated sub-clusters. This confirms our model successfully encodes discriminative temporal features within the relevant spatial context for both modalities.

    \section{Conclusion}
\label{sec:conclusion}

In this paper, we proposed \textbf{DETACH}, a novel decomposed spatio-temporal alignment framework for exocentric-

\begin{figure}[htbp]  
    \centering
    \includegraphics[width=0.47\textwidth]{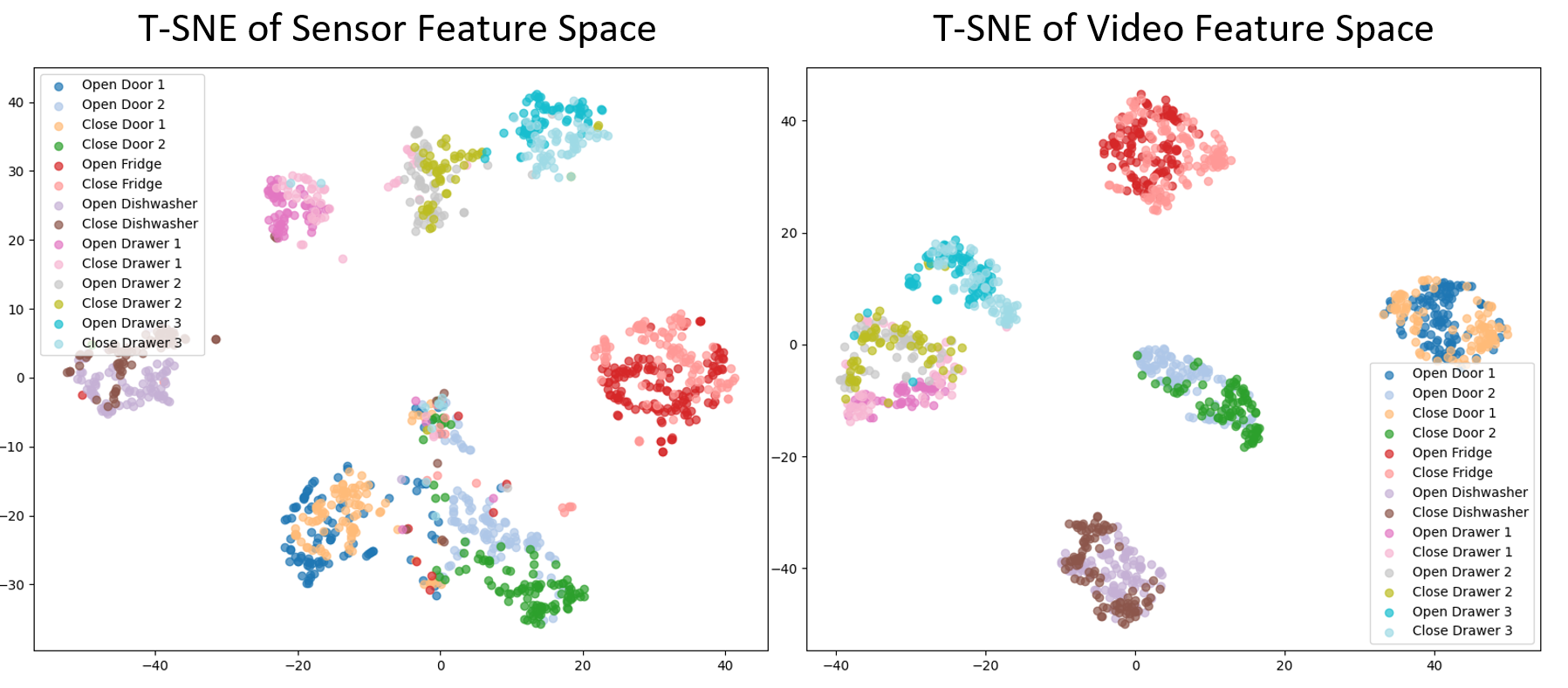}
    \vspace{-0.3cm}
    \caption{\textbf{t-SNE visualization of video and sensor feature spaces}. Samples are primarily grouped by spatial context at the global level, with fine-grained action-based sub-clusters emerging within each spatial group.}
    \label{fig:tsne}
\end{figure}

\noindent ambient settings. Our approach addresses the limitations of traditional Global Alignment by separating representations into spatial and temporal components. DETACH employs a two-stage strategy: Stage 1 establishes spatial grounding via online clustering and cross-modal refinement, while Stage 2 freezes spatial features to align temporal dynamics using our spatial-temporal weighted contrastive loss. This adaptive loss prioritizes hard negatives and mitigates false negatives, effectively learning discriminative temporal cues. Extensive experiments on Opportunity++ and HWU-USP datasets demonstrate that DETACH significantly outperforms existing unsupervised baselines. However, two limitations warrant future work. First, our current approach primarily focuses on sensors capturing explicit motion dynamics. The exploration of integrating various ambient sensors (e.g., ambient audio, light levels), which may provide valuable context, is left for future work. Second, our framework is designed for single-user scenarios due to the lack of available multi-user interaction datasets, which limits our ability to extend it to multi-user settings.

\clearpage

    {
        \small
        \putbib[reference]
    }
\end{bibunit}

 \begin{bibunit}[ieeenat_fullname]
     \clearpage
\setcounter{page}{1}
\maketitlesupplementary

\begin{strip}
    \centering
    \includegraphics[width=\textwidth]{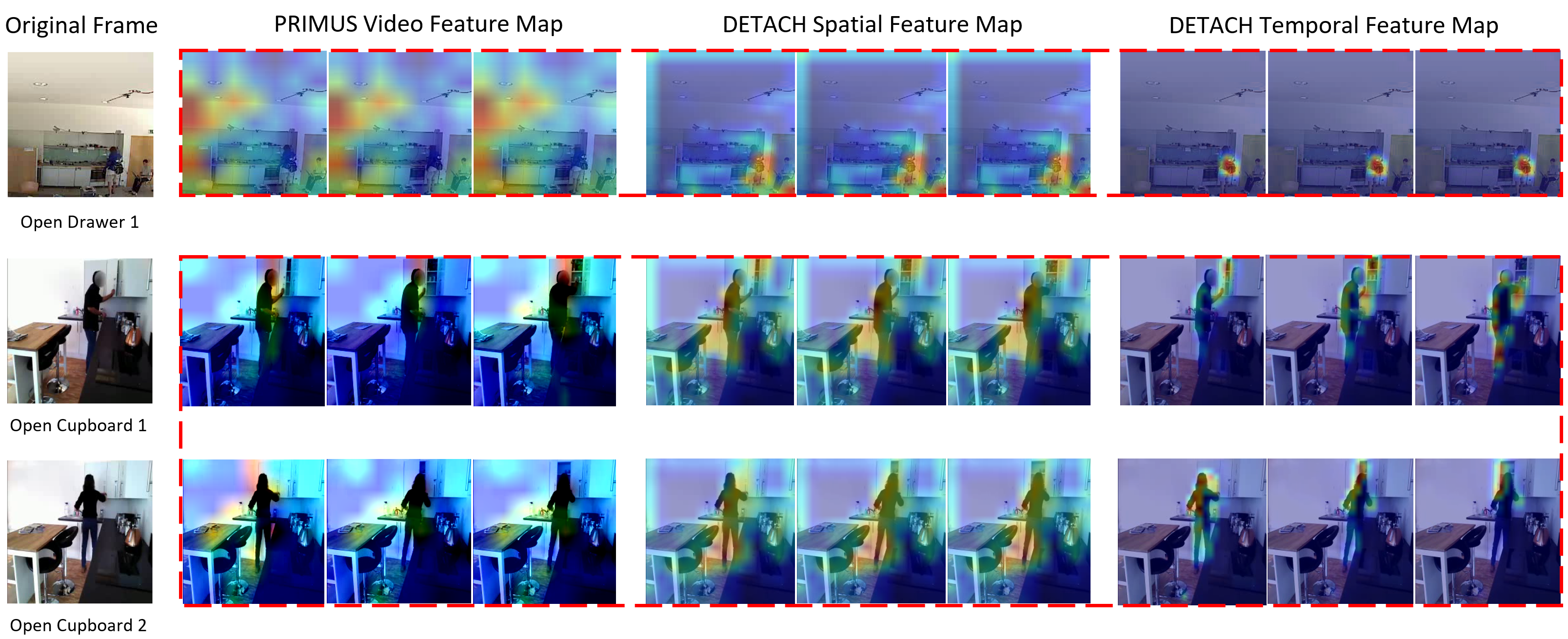}

    \captionsetup{hypcap=false}
    
    \captionof{figure}{\textbf{Qualitative visualization of decomposed spatial-temporal features.} We visualize the sequence of frames to illustrate the temporal flow. This visualization validates DETACH's effectiveness in two aspects: (1) capturing fine-grained subtle cues that are diluted in the baseline (Row 1), and (2) distinguishing similar actions by explicitly decomposing specific spatial contexts from shared temporal patterns (Row 2 and Row 3).}
    \label{fig:decompose_visualization}
\end{strip}


\section{Qualitative Analysis on Feature Decomposition}
Fig.~\ref{fig:decompose_visualization} visualizes the feature activation maps over the temporal frame sequence to qualitatively validate the effectiveness of DETACH in addressing the two fundamental limitations of Global Alignment discussed in the main text: (P1) the inability to capture local details and (P2) the over-reliance on modality-invariant temporal patterns.

Row 1 demonstrates the capability to capture fine-grained local details, specifically subtle motion patterns. As clearly observed in the baseline (PRIMUS \cite{das2025primus}) column, the attention maps are sparsely scattered across the entire frame without a clear focus. This visual evidence reflects the limitation of Global Alignment where compressing the video into a single vector causes minimal visual changes to be diluted against the dominant static scenes. In contrast, DETACH decomposes spatial context from temporal motion and thereby preserves these subtle cues. As a result, it precisely activates the specific regions corresponding to the interaction, effectively resolving the scattered attention issue of the baseline.

Row 2 and Row 3 compare two actions that share similar temporal patterns but involve different spatial contexts. The baseline feature maps for both rows focus primarily on dominant temporal patterns such as the human body rather than distinct spatial contexts. This observation visualizes the over-reliance on modality-invariant features which prevents the baseline from distinguishing actions with such similar dynamics. Conversely, DETACH successfully decomposes the representation. Its spatial feature map explicitly captures distinct spatial interaction location, while the temporal feature map independently handles the temporal dynamics. This decomposition effectively resolves the ambiguity that confounds the baseline model.

\begin{table*}[t]
\centering
\caption{\textbf{Robustness analysis across different sensor encoder architectures.} We report the performance using three different levels of sensor encoders on both datasets.}
\label{tab:encoder_robustness}
\resizebox{0.75\textwidth}{!}{%
\begin{tabular}{l|cc|cc}
\toprule
\multicolumn{1}{c|}{\multirow{2}{*}{\textbf{Sensor Encoder Architecture}}} & \multicolumn{2}{c|}{\textbf{Opportunity++}} & \multicolumn{2}{c}{\textbf{HWU-USP}} \\ \cmidrule(lr){2-3} \cmidrule(lr){4-5}
 & \textbf{F1 (Weighted)} & \textbf{mAP} & \textbf{F1 (Weighted)} & \textbf{mAP} \\ \midrule

\textbf{A. Lightweight} (1D CNN + GRU) & 0.62 & 0.73 & 0.71 & 0.67 \\ \midrule

\textbf{B. Heavyweight} (Transformer) & 0.71 & 0.85 & 0.71 & 0.66 \\ \midrule

\textbf{C. DETACH} (1D CNN + GRU + Attn.) & 0.73 & 0.83 & 0.71 & 0.67 \\ \midrule
\end{tabular}%
}
\end{table*}

\section{Impact of Sensor Encoder Architectures}
To investigate whether the performance improvements of DETACH stem from the proposed alignment framework or the specific encoder architecture, we evaluated our method using different sensor backbones:

\begin{itemize}
    \item \textbf{Lightweight:} A simple 1D-CNN followed by a GRU, representing previous egocentric multimodal studies \cite{das2025primus, moon2022imu2clip} with limited capacity.
    \item \textbf{Heavyweight:} A Transformer encoder with increased depth and parameters, designed to maximize representational capacity for multivariate sensor data, following recent approaches in time-series modeling \cite{zerveas2021transformer, xu2021limu}. 
    \item \textbf{DETACH:} Our default architecture, incorporating self-attention mechanism to capture spatio-temporal dependencies.
\end{itemize} 

As shown in Table~\ref{tab:encoder_robustness}, the Lightweight encoder significantly outperforms all state-of-the-art baselines reported in the main text, regardless of their backbone complexity. This result provides compelling evidence that the substantial performance gains are primarily driven by our novel decomposed spatio-temporal alignment strategy rather than model capacity. Furthermore, even when the heavyweight Transformer yields higher performance, the improvement over our default encoder is marginal. This suggests that our chosen architecture achieves an optimal balance between performance and efficiency, making it a highly effective and practical choice for ambient sensing environments without the computational cost of Transformers.

\section{Visualization of Feature Spaces on HWU-USP Dataset}
To complement the visualization provided in the main text for the Opportunity++ dataset (Fig. 7), we present the t-SNE visualization of the learned spatial features for the HWU-USP dataset. It is important to note the structural difference between the two datasets regarding label granularity. While Opportunity++ provides mid-level labels that allow for analyzing atomic actions within a specific context such as distinguishing opening a door from closing, HWU-USP dataset contains only high-level activity labels. Consequently, demonstrating the separation of atomic actions is not feasible for this dataset. 

Instead, Fig.~\ref{fig:tsne} visualizes the feature space at the spatial context level for both video and sensor modalities. The formation of cohesive clusters corresponding to underlying spatial contexts confirms that DETACH effectively learns discriminative spatial representations even when trained with coarse-grained high-level labels.

\begin{figure}[htbp]  
    \centering
    \includegraphics[width=0.47\textwidth]{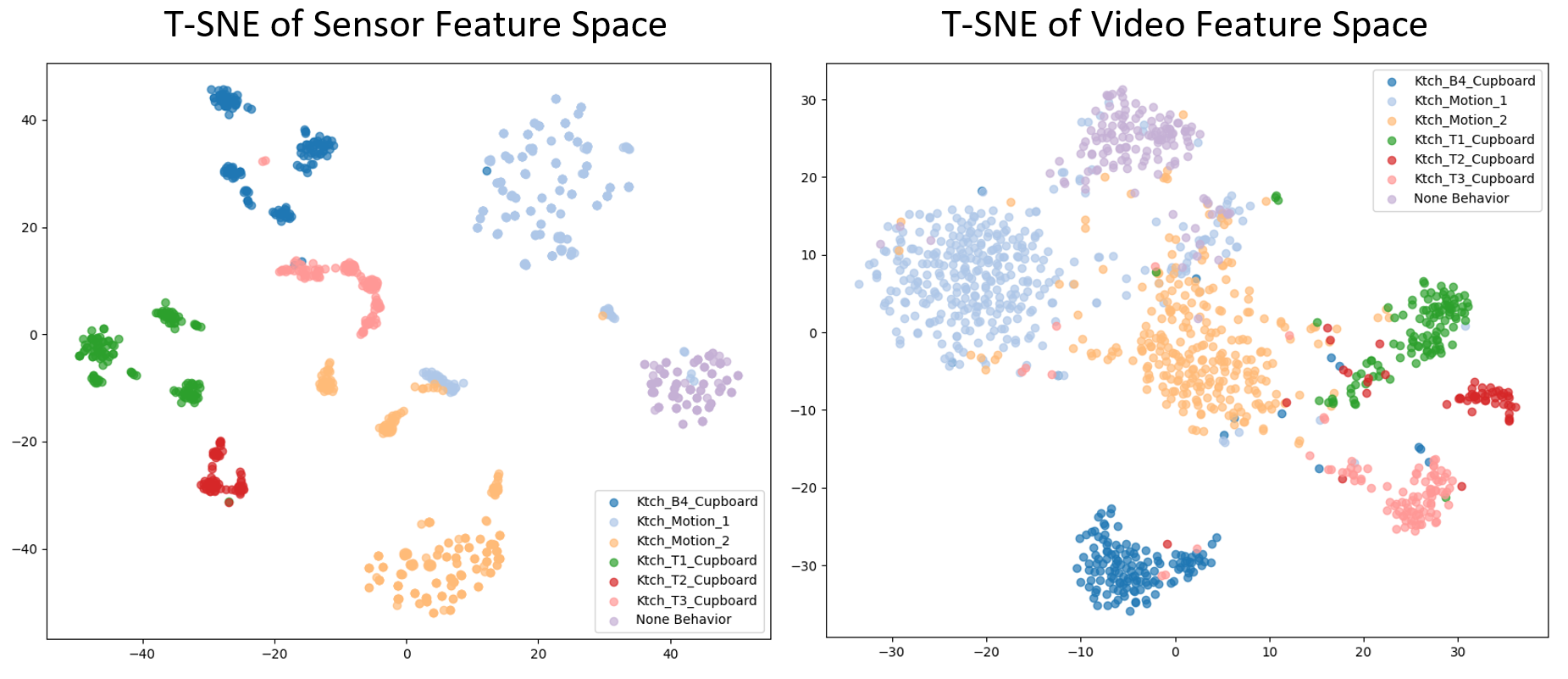}
    \vspace{-0.2cm}
    \caption{\textbf{t-SNE visualization of video and sensor feature spaces on HWU-USP Dataset}. Samples are grouped by spatial context for both modalities.}
    \label{fig:tsne}
\end{figure}

\section{Implementation Details}
To ensure reproducibility, we provide a detailed breakdown of the training strategy for DETACH. Our framework is trained in two sequential stages, each employing a scheduling strategy suited to its specific objective.

\noindent\textbf{Two-Stage Training Schedule.}
Since the convergence rate of online clustering varies with dataset distribution and complexity \cite{caron2018deep, caron2020unsupervised}, we adopted an adaptive training schedule for Stage 1 rather than a fixed epoch limit. We trained the spatial encoders until cluster assignments stabilized, which typically occurred around 25 epochs for Opportunity++ and 22 epochs for HWU-USP. These total epochs include the joint learning phase ($T=10$) described in the main text. This difference in convergence speed reflects the distinct complexities and distributions of the two datasets. 

For Stage 2, we used a fixed schedule of 50 epochs. As this stage focuses on optimizing our proposed Spatial-Temporal Weighted Contrastive Loss, a fixed budget ensures consistent evaluation of the alignment process, following standard evaluation protocols in contrastive learning \cite{chen2020simple}. The ``50 epochs'' mentioned in the main text and the ``Epochs'' axis (0, 20, 50) in Fig. 4 in the main text refer specifically to Stage 2. Thus, ``Epoch 0'' represents the state immediately after Stage 1 completion, where spatial features are already discriminative from Stage 1 convergence, while temporal features are initialized but not yet aligned.

\noindent\textbf{Stopping Criterion for Stage 1.}
To maintain strict adherence to the unsupervised protocol, we did not use ground-truth labels to determine when to stop Stage 1. Instead, we monitored the stability of cluster assignments by tracking the rate of change in pseudo-labels, defined as the fraction 

\begin{figure}[htbp]  
    \centering
    \includegraphics[width=0.47\textwidth]{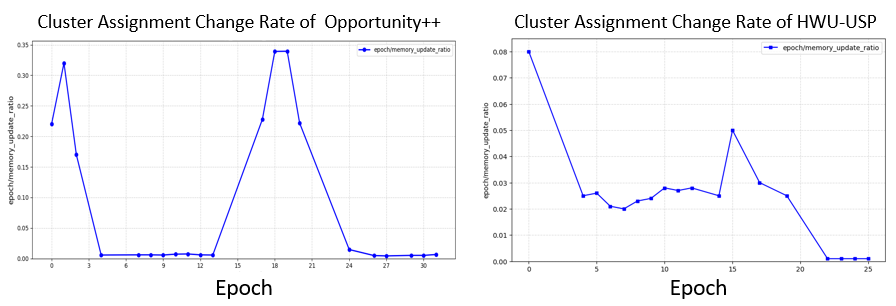}
    \vspace{-0.2cm}
    \caption{{\textbf{Stabilization of cluster assignments in Stage 1.} The pseudo-label change rate decreases and plateaus, indicating the convergence of the spatial encoder. The transient spike observed in the middle corresponds to the activation of the refinement loss ($L_{refine}$), which triggers a re-adjustment of cluster assignments. Subsequently, the rate stabilizes close to zero, serving as the stopping criterion for our unsupervised training protocol.}}
    \label{fig:cluster_assignment_rate}
\end{figure}

\noindent of samples that switched their predicted cluster assignments between consecutive epochs \cite{caron2018deep}. As shown in Fig.~\ref{fig:cluster_assignment_rate}, training was terminated when this rate reached a plateau close to zero, indicating that the model had learned discriminative spatial representations and the cluster assignments had converged.

\section{Dataset Details}

\noindent\textbf{Opportunity++.}
As detailed in Table~\ref{tab:dataset_details}, Opportunity++ \cite{ciliberto2021opportunity++} focuses on mid-level actions with short durations, such as ``Open Door 1'' or ``Close Fridge''. The dataset employs a hybrid sensor configuration of logic switches and accelerometers, requiring the model to align video representations with both discrete state transitions and continuous temporal dynamics. Data collected from a single subject provides a controlled environment for evaluating fine-grained alignment capabilities.

\noindent\textbf{HWU-USP.}
HWU-USP \cite{ranieri2021activity} targets high-level daily activities with longer temporal contexts, such as ``Making a Sandwich'' or ``Tidy Up''. With 16 different subjects, this dataset introduces intra-class variance in action execution. The sensor combines logic switches and logic motion sensors to capture object interactions and spatial occupancy. This multi-user setting evaluates the model's generalization ability and robustness in complex action recognition scenarios.

\begin{textblock}{80}(110, 25) 
    \centering
    \captionsetup{hypcap=false}
    \captionof{table}{\textbf{Detailed dataset statistics and configurations.}} \label{tab:dataset_details}
    \resizebox{\linewidth}{!}{%
        \begin{tabular}{l|c|c}
        \toprule
        \textbf{Attribute} & \textbf{Opportunity++} & \textbf{HWU-USP} \\ \midrule
        \multicolumn{3}{l}{\textit{\textbf{Sensor Configuration}}} \\ \midrule
        Sensor Types & \begin{tabular}[c]{@{}c@{}}Logic Switches ($\times 13$)\\ Accelerometers ($\times 7$)\end{tabular} & \begin{tabular}[c]{@{}c@{}}Logic Switches ($\times 4$)\\ Logic Motion Sensors($\times 2$)\end{tabular} \\ \midrule
        Sensor Location & \begin{tabular}[c]{@{}c@{}}Doors, Drawers,\\ Fridge, Dishwasher\end{tabular} & \begin{tabular}[c]{@{}c@{}}Cupboards, Wall\end{tabular} \\ \midrule
        \multicolumn{3}{l}{\textit{\textbf{Annotation Details}}} \\ \midrule
        Label Level & Mid-level & High-level \\ \midrule
        Total Classes & 14 & 5 \\ \midrule
        Class Names & \begin{tabular}[c]{@{}c@{}}Open/Close Door 1,2\\ Open/Close Drawer 1-3\\ Open/Close Fridge\\ Open/Close Dishwasher\end{tabular} & \begin{tabular}[c]{@{}c@{}}Cereals\\ Tea\\ Sandwich\\ Dishes\\ Tidy\end{tabular} \\ \midrule
        \multicolumn{3}{l}{\textit{\textbf{Experimental Setup}}} \\ \midrule
        Subjects & Single subject & 16 subjects (one per video) \\ \midrule
        Total Samples & 5324 windows & 3485 windows \\ \bottomrule
        \end{tabular}%
    }
\end{textblock}

\section{Additional Information}

We provide the complete source code and detailed experimental configurations to facilitate reproducibility. Further information regarding the implementation and training process is available at the following repository.

\vspace{0.5em} \noindent\href{https://github.com/anonymous-research2026/DETACH}{DETACH GitHub Repository}

\clearpage

    {
         \small
         \renewcommand{\refname}{References} 
         \putbib[reference]
    }
\end{bibunit}

\end{document}